\documentclass[11pt,a4paper]{article}
\usepackage[hyperref]{acl2018}
\usepackage{times}
\usepackage{latexsym}
\usepackage{url}
\usepackage{amstext}
\usepackage{amsmath}
\usepackage{forest}
\usepackage{qtree}
\usepackage{url}
\usepackage{mathtools}
\usepackage{multirow}
\usepackage{graphicx} 
\graphicspath{ {images/} }
\usepackage{tabularx}
\usepackage{ragged2e}
\newcolumntype{Y}{>{\RaggedRight\let\newline\\\arraybackslash\hspace{0pt}}X} 
\usepackage{tkz-berge}
\usepackage{tikz}
\usetikzlibrary{matrix,chains,positioning,decorations.pathreplacing,arrows}
\usepackage{tikz-qtree}
\usepackage{tikz-qtree-compat}
\usepackage[utf8]{inputenc}
\usepackage{xcolor}
\usepackage{pgfplots}
\usepackage{pgfpages}
\usepackage{subcaption}
\usepackage{array}
\usepackage{flushend}
\usepackage{arydshln}
\aclfinalcopy 
\def\aclpaperid{1607} 

\newcounter{Lcount}
\newcommand{\squishenum}{
	\begin{list}{\arabic{Lcount}. }
		{ \usecounter{Lcount}
			\setlength{\itemsep}{0pt}
			\setlength{\parsep}{0pt}
			\setlength{\topsep}{0pt}
			\setlength{\partopsep}{0pt}
			\setlength{\leftmargin}{2em}
			\setlength{\labelwidth}{1.5em}
			\setlength{\labelsep}{0.5em} } }
	
	\newcommand{\squishletter}{
		\begin{list}{\alph{Lcount}. }
			{ \usecounter{Lcount}
				\setlength{\itemsep}{0pt}
				\setlength{\parsep}{0pt}
				\setlength{\topsep}{0pt}
				\setlength{\partopsep}{0pt}
				\setlength{\leftmargin}{2em}
				\setlength{\labelwidth}{1.5em}
				\setlength{\labelsep}{0.5em} } }
		
		\newcommand{\squishlist}{
			\begin{list}{$\bullet$}
				{ \usecounter{Lcount}
					\setlength{\itemsep}{0pt}
					\setlength{\parsep}{0pt}
					\setlength{\topsep}{0pt}
					\setlength{\partopsep}{0pt}
					\setlength{\leftmargin}{2em}
					\setlength{\labelwidth}{1.5em}
					\setlength{\labelsep}{0.5em} } }

			\newcommand{\squishend}{
		\end{list} }
		\usepackage{pgf}
		\usetikzlibrary{arrows,decorations.pathmorphing,backgrounds,positioning,fit,petri,shapes.misc, arrows.meta,shapes.geometric,decorations.markings,calc,shadows.blur}
		\definecolor{myblue}{RGB}{0,128,255}
		\definecolor{myorange}{RGB}{211, 84, 0}
		\definecolor{mypurple}{RGB}{142, 68, 173}
		\definecolor{mygrey}{RGB}{158, 158, 158}
		\definecolor{lowpurple}{RGB}{204,153,255}
		\definecolor{lowwhite}{RGB}{255,255,255}
		\definecolor{verylowpurple}{RGB}{255,102,102}
		\definecolor{embcolor}{RGB}{255,255,255}
		\definecolor{myred}{RGB}{231, 76, 60}
		\definecolor{mygreen}{RGB}{162, 217, 206} 
		\definecolor{fontgrey}{RGB}{44, 62, 80}
		\definecolor{lowpurple}{RGB}{210, 180, 222}
		\definecolor{mypumpkin}{RGB}{229, 152, 102}
		\definecolor{lowgreen}{RGB}{171, 235, 198}
		\definecolor{lowred}{RGB}{245, 183, 177}
		\definecolor{pink}{RGB}{252,146,114}

\title{Learning Cross-lingual Distributed Logical Representations \\for Semantic Parsing}

\author{%
	Yanyan Zou \and Wei Lu\\
  Singapore University of Technology and Design \\
  8 Somapah Road, Singapore, 487372 \\
  {\tt yanyan\_zou@mymail.sutd.edu.sg, luwei@sutd.edu.sg} \\
}

\date{}

\begin{document}
\maketitle
\begin{abstract}


With the development of several multilingual datasets used for semantic parsing, recent research efforts have looked into the problem of learning semantic parsers in a multilingual setup \cite{duong2017multilingual,P17-2007}.
However, how to improve the performance of a monolingual semantic parser for a specific language by leveraging data annotated in different languages remains a research question that is under-explored.
In this work, we present a study to show how learning distributed representations of the logical forms from data annotated in different languages can be used for improving the performance of a monolingual semantic parser.
We extend two existing monolingual semantic parsers to incorporate such cross-lingual distributed logical representations as features.
Experiments show that our proposed approach is able to yield improved semantic parsing results on the standard multilingual GeoQuery dataset.


\end{abstract}

\section{Introduction}
Semantic parsing, one of the classic tasks in natural language processing (NLP), has been extensively studied in the past few years \cite{zettlemoyerlearning,WongYW:06,Wong:07,liang11dcs,kwiatkowski2011lexical:11,artzi2015broad:15}.
With the development of datasets annotated in different languages, learning semantic parsers from such multilingual datasets also attracted attention of researchers \cite{P17-2007}.
However, how to make use of such cross-lingual data to perform cross-lingual semantic parsing -- using data annotated for one language to help improve the performance of another language remains a research question that is largely under-explored.


\begin{figure}[tbp]
	\centering
	\resizebox{0.45\textwidth}{!}{\begin{tikzpicture}[level distance=30]
		\tikzset{level 1+/.style={sibling distance=0.4\baselineskip}}
		\Tree [.{Q{\small{UERY}} : \textsl{answer} (R{\small{IVER}})} [.{R{\small{IVER}}: \textsl{exclude} (R{\small{IVER}}, R{\small{IVER}})} {R{\small{IVER}} : \textsl{state} (all)} [.{R{\small{IVER}} : \textsl{traverse} (S{\small{TATE}})} [.{S{\small{TATE}} : \textsl{stateid} (S{\small{TATE}}N{\small{AME}})} {S{\small{TATE}}N{\small{AME}} : ($'texas'$)} ]] ]]
\end{tikzpicture} 
}}
\centering{\text{ English: which rivers do not run through texas ?}}
\centering{\text{ German: welche flüsse fliessen nicht durch texas ?}
\caption{An example of two semantically equivalent sentences (below) and their tree-shaped semantic representation (above).}
	\label{fig:ExampleForTreeAndSentence}
\end{figure}

Prior work \cite{Chan:05} shows that semantically equivalent words coming from different languages may contain shared semantic level information, which will be helpful for certain semantic processing tasks.
In this work, we propose a simple method to learn the distributed representations for output structured semantic representations which allow us to capture cross-lingual features.
Specifically, following previous work \cite{WongYW:06,Jones:12,Rhs:17}, we adopt a commonly used tree-shaped form as the underlying meaning representation where each tree node is a semantic unit.
Our objective is to learn for each semantic unit a distributed representation useful for semantic parsing, based on multilingual datasets.
Figure \ref{fig:ExampleForTreeAndSentence} depicts an instance of such tree-shaped semantic representations, which correspond to the two semantically equivalent sentences in English and German below it.

For such structured semantics, we consider each semantic unit separately.
We learn distributed representations for individual semantic unit based on multilingual datasets where semantic representations are annotated with different languages.
Such distributed representations capture shared information cross different languages.
We extend two existing monolingual semantic parsers \cite{Luw:15,Rhs:17} to incorporate such cross-lingual features.
To the best of our knowledge, this is the first work that exploits cross-lingual embeddings for logical representations for semantic parsing.
Our system is publicly available at {\url{http://statnlp.org/research/sp/}}.

\section{Related Work}
Many research efforts on semantic parsing have been made, such as mapping sentences into lambda calculus forms based on CCG \cite{yoav11boot,yoav14compact,kwiatkowski2011lexical:11}, modeling dependency-based compositional semantics \cite{liang11dcs,liang17dcs}, or transforming sentences into tree structured semantic representations \cite{Luw:15,Rhs:17}.
With the development of multilingual datasets, systems for multilingual semantic parsing are also developed.
\newcite{ZMJie:14} employed majority voting to combine outputs from different parsers for certain languages to perform multilingual semantic parsing.
\newcite{P17-2007} presented an extension of one existing neural parser, {\textsc{Seq2Tree}} \cite{P16-1004}, by developing a shared attention mechanism for different languages to conduct multilingual semantic parsing.
Such a model allows two types of input signals: single source {SL-{\textsc{Single}} and multi-source  {SL-{\textsc{Multi}}}.
However, semantic parsing with cross-lingual features has not been explored, while many recent works in various NLP tasks show the effectiveness of shared information cross different languages. Examples include semantic role labeling \cite{kozhevnikov13crosssrl}, information extraction \cite{jeff13transfer,heng17tagging,radu17ner}, and question answering \cite{israa17qa}, which motivate this work.
%

{Our work involves exploiting distributed output representations for improved structured predictions, which is in line with works of \cite{NIPS2014_5323,W14-2409,xiaoguo15adaption}.
The work of \cite{W14-2409} is perhaps the most related to this research.
The authors first map first-order logical statements produced by a semantic parser or an information extraction system into expressions in tensor calculus.
They then learn low-dimensional embeddings of such statements with the help of a given logical knowledge base consisting of first-order rules so that the learned representations are consistent with these rules.
They adopt stochastic gradient descent (SGD) to conduct optimizations. This work learns distributed representations of logical forms from cross-lingual data based on co-occurrence information without relying on external knowledge bases.
}

\section{Approach}

\subsection{Semantic Parser}
\label{sec:nht}

In this work, we build our model and conduct experiments on top of the discriminative hybrid tree semantic parser \cite{Luw:14,Luw:15}.
The parser was designed based on the hybrid tree representation ({\textsc{HT-g}}) originally introduced in \cite{Luw:08}.
The hybrid tree is a joint representation encoding both sentence and semantics that aims to capture the interactions between words and semantic units.
A discriminative hybrid tree ({\textsc{HT-d}}) \cite{Luw:14,Luw:15} learns the optimal latent word-semantics correspondence where every word in the input sentence is associated with a semantic unit.
Such a model allows us to incorporate rich features and long-range dependencies.
Recently, \newcite{Rhs:17} extended {\textsc{HT-d}} by attaching neural architectures, resulting in their neural hybrid tree ({\textsc{HT-d} \textsc{(nn)}}). 

Since the correct correspondence between words and semantics is not explicitly given in the training data, we regard the hybrid tree representation as a latent variable.
Formally, for each sentence $\mathbf{n}$ with its semantic representation $\mathbf{m}$ from the training set, we assume the joint representation (a hybrid tree) is $\mathbf{h}$.
Now we can define a discriminative log-linear model as follows:
{\begin{gather}
\small
P_{\Lambda}(\mathbf{m}|\mathbf{n}) =\sum_{\mathbf{h} \in \mathcal{H}{(\mathbf{n},\mathbf{m})}} P_{\Lambda}(\mathbf{m},\mathbf{h}|\mathbf{n})   \notag \\
=\frac{\sum_{\mathbf{h} \in \mathcal{H}{(\mathbf{n},\mathbf{m})}} e^{F_{\Lambda}(\mathbf{n},\mathbf{m},\mathbf{h})}}{\sum_{\mathbf{m}',\mathbf{h}'\in \mathcal{H}(\mathbf{n},\mathbf{m}')}e^{F_{\Lambda}(\mathbf{n},\mathbf{m}',\mathbf{h}'))}}
\\
F_{\Lambda}(\mathbf{n},\mathbf{m},\mathbf{h})) = \Lambda\cdot\Phi{(\mathbf{n},\mathbf{m},\mathbf{h}))}
\label{partition}
\end{gather}
where $\mathcal{H}(\mathbf{n},\mathbf{m})$ returns the set of all possible joint representations that contain both $\mathbf{n}$ and $\mathbf{m}$ exactly, and $F$ is a scoring function that is calculated as a dot product between a feature function $\Phi$ defined over tuple ($\mathbf{m}$, $\mathbf{n}$, $\mathbf{h}$) and a  weight vector $\Lambda$. 


To incorporate neural features, {\textsc{HT-d} \textsc{(nn)}} defines the following scoring function:
\begin{gather}
F_{\Lambda, \Theta}(\mathbf{n},\mathbf{m},\mathbf{h})) = \Lambda\cdot\Phi{(\mathbf{n},\mathbf{m},\mathbf{h})} + G_{\Theta}(\mathbf{n},\mathbf{m},\mathbf{h})
\label{scof1}
\end{gather} 
where $\Theta$ is the set of parameters of the neural networks and $G$ is the neural scoring function over the ($\mathbf{n}$,$\mathbf{m}$,$\mathbf{h}$) tuple \cite{Rhs:17}.
Specifically, the neural features are defined over a fixed-size window surrounding a word in $\mathbf{n}$ paired with its immediately associated semantic unit. 
Following the work \cite{Rhs:17}, we denote the window size as $J\in\{0,1,2\}$.

\subsection{Cross-lingual Distributed Semantic Representations}
\label{sec:semantic_embedding}


A multilingual dataset used in semantic parsing comes with instances consisting of logical forms annotated with sentences from multiple different languages.
In this work, we aim to learn one monolingual semantic parser for each language, while leveraging useful information that can be extracted from other languages.
Our setup is as follows. Each time, we train the parser for a target language and regard the other languages as {\em auxiliary languages}.
To learn cross-lingual distributed semantic representations from such data,
we first combine all data involving all auxiliary languages to form a large dataset.
Next,
for each target language, we  construct a semantics-word co-occurrence matrix ${\mathbf{M}}\in{R}^{m\times n}$ (where $m$ is the number of unique semantic units, $n$ is the number of unique words in the combined dataset).
Each entry is the number of co-occurrences for a particular (semantic unit-word) pair.
We will use this matrix to learn a low-dimensional cross-lingual representation for each semantic unit.
To do so, we first apply singular value decomposition (SVD) to this matrix, leading to:
\begin{gather}
\mathbf{M} = \mathbf{U}\mathbf{\Sigma}\mathbf{V^\ast}
\end{gather}
where ${\mathbf{U}}\in{R}^{m\times m}$ and ${\mathbf{V}}\in{R}^{n\times m}$ are unitary matrices, ${\mathbf V}^\ast$ is the conjugate transpose of ${\mathbf V}$, and $\mathbf{\Sigma}\in R^{m\times m}$ is a diagonal matrix.
We truncate the diagonal matrix ${\mathbf{\Sigma}}$ and left multiply it with $\mathbf{U}$:
\begin{gather}
\mathbf{U}\mathbf{\tilde{\Sigma}}
\end{gather}
where $\mathbf{\tilde{\Sigma}}\in R^{m\times d}$ is a matrix that consists of only the left $d$ columns of $\mathbf{\Sigma}$, containing the $d$ largest singular values.
We leave the rank $d$ as a hyperparameter.
Each row in the above matrix is a $d$-dimensional vector, giving a low-dimensional representation for one semantic unit.
Such distributed output representations can be readily used as continuous features in $\Phi{(\mathbf{n},\mathbf{m},\mathbf{h})}$.

\subsection{Training and Decoding}

During the training process, we optimize the objective function defined over the training set as:
\begin{gather}
\mathcal{L}(\Lambda,\Theta)=\sum_{i}\log \sum_{\mathbf{h}\in\mathcal{H}(\mathbf{n}_i,\mathbf{m}_i) }e^{F_{\Lambda,\Theta}(\mathbf{n}_i,\mathbf{m}_i,\mathbf{h})} \notag \\
- \sum_{i}\log \sum_{\mathbf{m}',\mathbf{h}'\in\mathcal{H}(\mathbf{n}_i,\mathbf{m}') }e^{F_{\Lambda,\Theta}(\mathbf{n}_i,\mathbf{m}',\mathbf{h}')}
\label{obj6}
\end{gather} 
We follow the dynamic programming approach used in \cite{Rhs:17} to perform efficient inference, and follow the same optimization strategy as described there.

In the decoding phase, we are given a new input sentence $\mathbf{n}$, and find the optimal semantic tree $\mathbf{m}^{\ast}$:
\begin{align}
\mathbf{m}^{\ast} 
&=\mathop{\arg\max}_{\mathbf{m,h}\in \mathcal{H}(\mathbf{n},\mathbf{m})}{F_{\Lambda, \Theta}(\mathbf{n},\mathbf{m},\mathbf{h})}
\label{argmax}
\end{align}

\begin{table}[t]
	\centering
	\small
	\resizebox{0.48\textwidth}{!}{\begin{tabular}{|l|cc||l|cc|}
		\hline
		& Rank ($d$)  & $F$ &  & Rank ($d$)  & $F$ \\
		\hline
		English & 30 & 88.3 &Chinese & 10 & 80.0  \\
		Thai    & 20 & 85.8 &Indonesian  & 30  & 88.3\\
		German  & 30 & 78.3 &Swedish  & 20 & 83.3   \\
		Greek  	& 10 & 81.7 &Farsi  & 10 &  76.7\\
		\hline
	\end{tabular}}
	\caption{Performance on development set. $F$: F1-measure (\%).}
	\label{tab:hyperparameter}
\end{table}

\begin{table*}[t!]
	\centering
	\resizebox{\textwidth}{!}{\begin{tabular}{|ll|cc|cc|cc|cc|cc|cc|cc|cc|}
			\hline
			\multicolumn{2}{|c|}{\multirow{2}{*}{}} &
			\multicolumn{2}{c|}{English} &
			\multicolumn{2}{c|}{Thai} & 
			\multicolumn{2}{c|}{German} &
			\multicolumn{2}{c|}{Greek}&
			\multicolumn{2}{c|}{Chinese} &
			\multicolumn{2}{c|}{Indonesian} & 
			\multicolumn{2}{c|}{Swedish} &
			\multicolumn{2}{c|}{Farsi} \\
			&&  $Acc.$ & $F$ & $Acc.$ & $F$ & $Acc.$ & $F$ & $Acc.$ & $F$ & $Acc.$ & $F$ & $Acc.$ & $F$ & $Acc.$ & $F$ & $Acc.$ & $F$ \\
			\hline
			\hline
			{\textsc{Wasp}} && 71.1&77.7 & 71.4  &  75.0& 65.7&74.9 & 70.7  & 78.6& 48.2&51.6 & 74.6  &  {\bf 79.8}& 63.9&71.5 & 46.8 & 54.1\\
			{\textsc{HT-g}}   && 76.8&81.0 & 73.6  &  76.7& 62.1&68.5 & 69.3  & 74.6 & 56.1&58.4 & 66.4  &  72.8& 61.4&70.5 & 51.8  & 58.6 \\
			\textsc{UBL-s}& & 82.1 & 82.1 &66.4& 66.4 &75.0 &75.0 &73.6 &73.7&63.8&63.8&73.8&73.8&78.1&78.1&64.4&64.4\\
            {\textsc{TreeTrans}}& & 79.3 & 79.3 &78.2 & 78.2 &74.6 &74.6 &75.4 &75.4&-&-&-&-&-&-&-&-\\
			{\textsc{Seq2Tree}}$\dagger$&& 84.5&-&71.9&-&70.3&-&73.1&-&73.3&-&{\bf 80.7}&-&{\bf 80.8} &-&70.5&-\\
			SL-{\textsc{Single}} $\dagger$&&83.5&-&72.1&-&69.3&-&74.2&-&74.9&-&79.8&-&77.5&-&72.2&-\\
			\hline
			\hline
			\textsc{HT-d}& & {\bf 86.8} &  {\bf 86.8} &{80.7}  &80.7 &{\bf 75.7} &{\bf 75.7} &79.3 &79.3&76.1&76.1&75.0&75.0&{79.3}&{\bf 79.3}&73.9&73.9\\
			\hdashline
			\textsc{HT-d} (+\textsc{o})& &86.1 &86.1&{\bf 81.1} &{\bf 81.1} & 73.6&73.6 &{\bf 81.4} &{\bf 81.4}&{\bf 77.9}&{\bf 77.9}&{79.6}&{79.6}&{79.3}&{\bf 79.3}&{\bf 75.7}&{\bf 75.7}\\				
			\hline
			\hline
			\multirow{3}{*}{\textsc{HT-d} (\textsc{nn})}  &$J$=0 &  87.9&87.9  &82.1 &82.1 &75.7&75.7 &81.1 &81.1&76.8&76.8&76.1&76.1&81.1&81.1&75.0&75.0\\
			&$J$=1 &88.6  &88.6 &84.6&84.6  &\textbf{76.8} &\textbf{76.8} &79.6 &79.6&75.4&75.4&78.6&78.6&82.9&82.9&76.1&76.1 \\
			&$J$=2 & \textbf{90.0} &\textbf{90.0}  &  82.1&82.1  & 73.9&73.9 &80.7 &80.7&81.1&81.1&{81.8}&{81.8}&{\bf83.9}&{\bf83.9}&74.6&74.6\\
			\hdashline
			\multirow{3}{*}{\textsc{HT-d} (\textsc{nn}+\textsc{o})} &$J$=0 &{86.1}  &{86.1}   & {83.6} &{83.6}   &{73.9} &{73.9}& 82.1& 82.1&77.9&77.9&81.1&81.1&82.1&82.1&74.6&74.6\\
			&$J$=1 &86.1 &86.1  &\textbf{86.1}   & \textbf{86.1}   &72.5 &72.5 &80.4 & 80.4&{81.4}&{81.4}&{82.5}&{82.5}&82.5&82.5&75.7&75.7\\
			&$J$=2 &{89.6}  &{89.6}   &{84.6}  & {84.6}   & {72.1}&{72.1}&\textbf{83.2} &\textbf{83.2}& \textbf{82.1} &\textbf{82.1}& \textbf{83.9}&\textbf{83.9} &{83.6}&{83.6}&\textbf{76.8}&\textbf{76.8}\\
			\hline
		\end{tabular}
	}
	\caption{Performance on multilingual datasets. $Acc.$: accuracy (\%), $F$: F1-measure (\%). +\textsc{o}: including distributed representations for semantic units as features. ($\dagger$ indicates systems that make use of lambda calculus expressions as meaning representations.)}
	\label{tab:model_comparison}
\end{table*}

Again, the above equation can be efficiently computed by dynamic programming \cite{Rhs:17}.
\label{sec:exp}

\section{Experiments and Results}

\subsection{Datasets and Settings}

We evaluate our approach on the standard GeoQuery dataset annotated in eight languages \cite{WongYW:06,Jones:12,Rhs:17}.
We follow a standard practice for evaluations which has been adopted in the literature \cite{Luw:14,Luw:15,Rhs:17}.
Specifically, to evaluate the proposed model, predicted outputs are transformed into Prolog queries.
An output is considered as correct if answers that queries retrieve from GeoQuery database are the same as the gold ones .
The dataset consists of 880 instances.
In all experiments, we follow the experimental settings and procedures in \cite{Luw:14,Luw:15,Rhs:17}.
In particular, we use 600 instances for training and 280 for test and set the maximum optimization iteration to 150.
In order to tune the rank $d$, we randomly select 80\% of the training instances for learning and use the rest 20\% for development.
We report the value of $d$ for each language in Table \ref{tab:hyperparameter} and the F1 scores on the development set.
%
%
%

\subsection{Results}
We compare our models against different existing systems, especially the two baselines \textsc{HT-d} \cite{Luw:15} and \textsc{HT-d} \textsc{(nn)} \cite{Rhs:17} with different word window sizes $J \in \{0,1,2\}$.

{{\textsc{Wasp}} \cite{WongYW:06} is a semantic parser based on statistical phrase-based machine translation.
{{\textsc{UBL-s}}} {\cite{kwiatkowski2010inducing:10}} induced probabilistic CCG grammars with higher-order unification that allowed to construct general logical forms for input sentences.
{T{\small{REE}}T{\small{RANS}}} \cite{Jones:12} is built based on a Bayesian inference framework.
We run {{\textsc{Wasp}}}, {{\textsc{UBL-s}}}, {\textsc{HT-g}}, {\textsc{UBL-s}}, \textsc{Seq2Tree} and \textsc{SL-Single} \footnote{
	Note that in \newcite{P16-1004}, they adopted a different split -- using 680 instances as training examples and the rest 200 for evaluation.
	We ran the released source code for \textsc{Seq2Tree} \cite{P16-1004} over eight different languages.
	For each language, we repeated the experiments 3 times with different random seed values, and reported the average results as shown in Table \ref{tab:model_comparison} to make comparisons.
	Likewise, we ran \textsc{SL-Single} following the same procedure.} for comparisons.
Note that there exist multiple versions of logical representations used in the \textsc{GeoQuery} dataset.
Specifically, one version is based on lambda calculus expression, and the other is based on the variable free tree-shaped representation.
We use the latter representation in this work, while the {\textsc{Seq2Tree}} and {SL-{\textsc{Single}} employ the lambda calculus expression.
It was noted in \newcite{kwiatkowski2010inducing:10,Luw:14} that evaluations based on these two versions are not directly comparable -- the version that uses tree-shaped representations appears to be more challenging.
{We do not compare against \cite{ZMJie:14} due to their different setup from ours.\footnote{
They trained monolingual semantic parsers. In the evaluation phase, they fed parallel text from different languages to each individual semantic parser, then employed a majority voting based ensemble method to combine predictions. Differently, we train monolingual semantic parsers augmented with cross-lingual distributed semantic information. In the evaluation phase, we only have one monolingual semantic parser. Hence, these two efforts are not directly comparable.
}
\begin{figure*}[t]
	\centering
	\scalebox{1.02}{
\includegraphics[width=13.0cm,height=6.8cm]{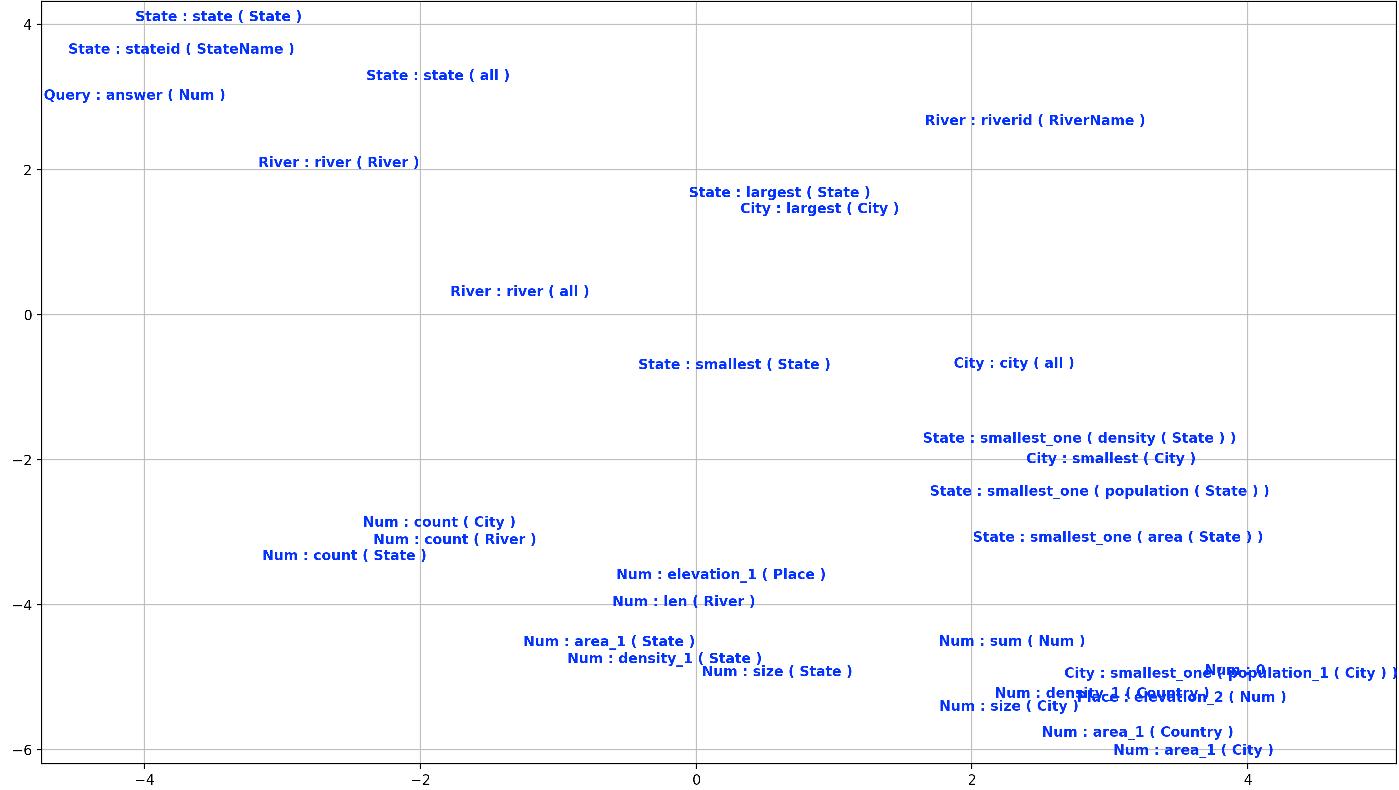}
}
\caption{2-D projection of learned distributed representations for semantics.}
	\label{fig:embedding}
	\vspace{-4mm}
\end{figure*}

Table \ref{tab:model_comparison} shows results that we have conducted on eight different languages.
The highest scores are highlighted.
We can observe that when distributed logical representations are included, both \textsc{HT-d} and \textsc{HT-d} (\textsc{nn}) can lead to competitive results.
Specifically, when such features are included, evaluation results for 5 out of 8 languages get improved.

\textcolor{black}{We found that the shared information cross different languages could guide the model so that it can make more accurate predictions,
eliminating certain semantic level ambiguities associated with the semantic units.
This is exemplified by a real instance from the English portion of the dataset:}
\begin{table}[h]
	\centering
	\scalebox{0.86}
	{
		
		\begin{tabular}{ll}
			{Input}:& Which states have a river? \\
			{Gold}:& $answer(state(loc(river(all))))$ \\
			\hline
			{Output}:& $answer(state(traverse(river(all))))$ \\
			{Output \textsc{(+o)}}:& $answer(state(loc(river(all))))$ \\
		\end{tabular}
	}
	\label{tab:problem_definition}
\end{table}

\textcolor{black}{Here  the input sentence in English is ``Which states have a river?", and the correct output is shown below the sentence.
Output is the prediction from \textsc{HT-d (nn)} and Output (\textsc{+o}) is the parsing result given by \textsc{HT-d (nn+o)} where the learned cross-lingual representations of semantics are included.
We observe that, by introducing our learned cross-lingual semantic information, the system is able to distinguish the two semantically related concepts, ${loc}$ (located in) and ${traverse}$ (traverse), and further make more promising predictions.}

Interestingly, for German, the results become much lower when such features are included,
indicating such features are not helpful in the learning process when such a language is considered.
Reasons for this need further investigations. We note, however, previously it was also reported in the literature that the behavior of the performance associated with this language is different than other languages in the presence of additional features \cite{Luw:14}.

\subsection{Visualizing Output Representations}

To qualitatively understand how good the learned distributed representations are, we also visualize the learned distributed representations for semantic units.
In the Figure \ref{fig:embedding}, we plot the embeddings of a small set of semantic units which are learned from all languages other than English.
Each representation is a 30-dimensional vector and is projected into a 2-dimensional space using Barnes-Hut-SNE \cite{maaten2013barnes} for visualization.
In general, we found that semantic units expressing similar meanings tend to appear together. For example, the two semantic units {S{\small{TATE}} : $smallest\_one$ ( $density$ (S{\small{TATE}}))} and {S{\small{TATE}} : $smallest\_one$ ( $population$ (S{\small{TATE}}))} share similar representations.
However, we also found that occasionally semantic units conveying opposite meanings are also grouped together.
This reveals the limitations associated with such a simple co-occurrence based approach for learning distributed representations for logical expressions.

\section{Conclusions}

In this paper, we empirically show that the distributed representations of logical expressions learned from multilingual datasets for semantic parsing can be exploited to improve the performance of a monolingual semantic parser.
Our approach is simple, relying on an SVD over semantics-word co-occurrence matrix for finding such distributed representations for semantic units.
Future directions include investigating better ways of learning such distributed representations as well as learning such distributed representations and semantic parsers in a joint manner.

\section*{Acknowledgments}
We would like to thank the three anonymous ACL reviewers for their thoughtful and constructive comments.
We would also like to thank Raymond H. Susanto for his help on this work.
This work is supported by Singapore Ministry of Education Academic Research Fund (AcRF) Tier 2 Project MOE2017-T2-1-156, and is partially supported by Singapore Ministry of Education Academic Research Fund (AcRF) Tier 1 SUTDT12015008.
\bibliography{semantic_embedding}
\bibliographystyle{acl_natbib}

%
%

\end{document}